\documentclass[10pt,twocolumn,letterpaper]{article}

\usepackage[pagenumbers]{cvpr} 

\usepackage{graphicx}
\usepackage{amsmath,amssymb}
\usepackage{booktabs}
\usepackage{multirow}
\usepackage{microtype}
\usepackage{xcolor}
\usepackage{colortbl}
\usepackage{pifont}
\newcommand{\cmark}{\textcolor{green!50!black}{\ding{51}}}
\newcommand{\xmark}{\textcolor{red!75!black}{\ding{55}}}
\newcommand{\otherref}[2]{#2}

\newcommand{\method}{\textsc{Reveal}}
\newcommand{\methodself}{\method{} (Self-Depth)}
\newcommand{\methodda}{\method{} (DA3)}

\renewcommand{\topfraction}{0.95}
\renewcommand{\dbltopfraction}{0.95}
\renewcommand{\bottomfraction}{0.6}
\renewcommand{\textfraction}{0.05}
\renewcommand{\floatpagefraction}{0.7}
\renewcommand{\dblfloatpagefraction}{0.7}

\definecolor{cvprblue}{rgb}{0.21,0.49,0.74}
\usepackage[pagebackref,breaklinks,colorlinks,allcolors=cvprblue]{hyperref}

\def\confName{CVPR}
\def\confYear{2027}

\title{Complementary Geometry in Gaussian Splatting}

\author{Mingle Jiang$^{1}$, Jikui Zhou$^{1}$, Chenghan Lyu$^{1}$, Rui Xu$^{2}$, Yunke Wang$^{1}$, Chang Xu$^{1}$\\
\textnormal{$^{1}$The University of Sydney}\quad
\textnormal{$^{2}$CCNU}
}

\begin{document}
\maketitle


\begin{abstract}
Gaussian Splatting has achieved high visual quality in novel view synthesis, but the accuracy of its rendered geometry remains a challenge. In particular, the rendered depth is computed as an opacity-weighted average along each ray, which may not correspond to the actual scene surface. In this paper, we observe that depth predictions from different views of the same frozen Gaussian representation exhibit complementary errors, allowing us to improve novel-view geometry using information from other views without modifying the original representation. 
Based on this observation, we propose \method{}, a general framework that selectively refines novel-view depth using complementary geometric information. 
Specifically, we first reproject depth predictions from source views to the target view and combine them using robust multi-view consensus. We then introduce a learned selection mechanism to determine whether the candidate depth is more accurate than the original prediction, based on the expected reduction in depth error. In addition, we introduce RIG, a real-world benchmark captured with synchronized RGB-D cameras, which provides ground-truth depth for evaluating both static and dynamic reconstruction models. Experiments on 6 Gaussian Splatting backbones across 3 benchmarks show that \method{} consistently improves depth accuracy while maintaining comparable rendering quality.

\end{abstract}


\section{Introduction}
\label{sec:intro}

\begin{figure}[t]
  \centering
  \suppressfloats[t]
  \includegraphics[width=\linewidth]{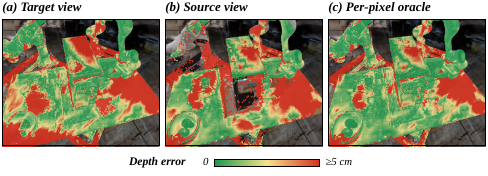}
  \caption{Complementary depth errors across the novel view. (a) Depth error from direct target-view rendering. (b) Depth error from source-view rendering, reprojected into the target view. (c) Per-pixel oracle selecting the depth estimate with lower ground-truth error. The complementary error patterns in (a) and (b) allow the oracle to substantially reduce depth error without modifying the underlying representation.}
  \vskip -0.1in
  \label{fig:motivation}
\end{figure}

\begin{figure*}[!t]
  \centering
  \includegraphics[width=\textwidth]{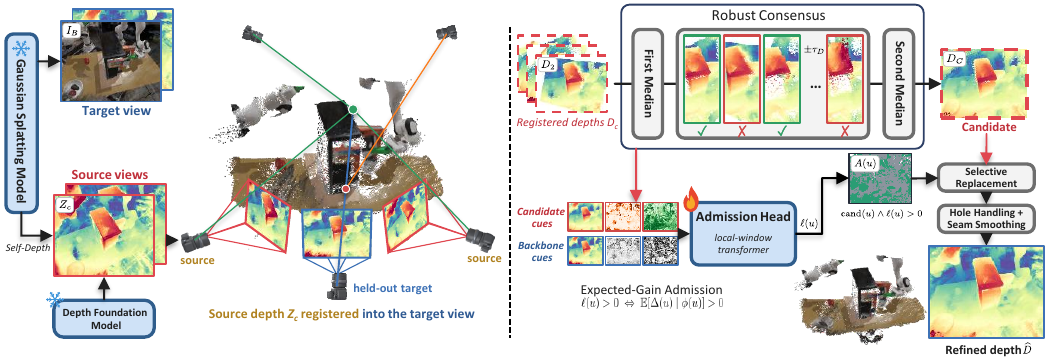}
  \caption{Overview of \method{}.
\textbf{Left:} The frozen Gaussian Splatting model renders target-view color and depth. Source-view depth is obtained either from the backbone itself or from a frozen depth foundation model, and then reprojected into the target view using known camera poses. \textbf{Right:} Robust multi-view consensus combines the reprojected depths to construct a candidate depth map. An expected-gain admission head compares the candidate with the backbone prediction and selects corrections expected to reduce depth error. The selected corrections are applied to refine novel-view depth without modifying the original Gaussian representation.}
\label{fig:schematic}
\end{figure*}

Gaussian Splatting (GS) \citep{chen2024mvsplat,liu2025monosplat,zhang2025transplat,wu2024four,wu2025swift4d,zhou2026c4d}  has achieved impressive results in novel view synthesis, enabling high-quality image rendering from previously unobserved viewpoints \citep{kerbl2023gaussian}. 
Despite these advances, most methods are primarily evaluated by the visual quality of rendered images, using metrics such as PSNR. However, high visual quality does not necessarily indicate accurate geometry. In particular, Gaussian Splatting typically computes rendered depth as an opacity-weighted average along each camera ray, which may not correspond to the actual scene surface \citep{chen2025pgsr,huang2024twodgs}. As a result, a model can produce visually convincing novel views while introducing errors in object depth and spatial relationships. Such errors can limit the use of reconstructed scenes in applications that require accurate 3D geometry, such as robotic manipulation and spatial reasoning.

We observe that depth predictions rendered from different views of the same Gaussian representation exhibit complementary errors. As shown in Figure~\ref{fig:motivation}, depth rendered directly at a target view and depth rendered from a source view and reprojected to the target view can have different error patterns. Although neither prediction is consistently more accurate, one can provide better depth estimates at locations where the other is inaccurate. This observation suggests that depth predictions from other views can be used to improve novel-view geometry without modifying the underlying Gaussian representation.


Based on this observation, we propose \method{}, a general framework that selectively refines novel-view geometry using complementary depth predictions. As shown in Figure~\ref{fig:schematic}, we first construct candidate geometry by reprojecting and aggregating depth predictions from source views. However, complementary predictions do not always provide better geometry, and directly combining them may introduce additional errors. We therefore formulate geometric refinement as a selective correction problem, where each replacement is determined by its expected reduction in depth error. To this end, we introduce a learned admission mechanism that compares the candidate geometry with the original prediction and selects corrections that are expected to improve accuracy. This allows \method{} to exploit complementary geometric information while retaining reliable predictions from the frozen backbone.



To evaluate geometric accuracy under real-world conditions, we further introduce RIG, a tabletop benchmark captured using three synchronized RGB-D cameras. RIG contains 130 recorded takes of object manipulation, including 95 evaluation takes with 2,280 held-out target views per backbone. By using measured depth from a held-out camera as ground truth, RIG provides a common evaluation protocol for both static and dynamic reconstruction models under real-world motion and occlusion. We evaluate \method{} on six Gaussian Splatting backbones across DTU \citep{jensen2014dtu}, LBM \citep{tri2025lbm} and our benchmark RIG. The results show consistent improvements in novel-view depth accuracy while preserving rendering quality.

Our main contributions are summarized as follows:
\begin{itemize}
    \item We identify complementary depth errors across different views of the same frozen Gaussian representation and show that these errors can be exploited to improve novel-view geometry.
    \item We propose \method{}, a backbone-agnostic framework that combines robust multi-view consensus with learned selection based on expected depth-error reduction.
    \item We introduce RIG, a real-world RGB-D benchmark for evaluating novel-view geometry, and demonstrate consistent improvements across six static and dynamic Gaussian Splatting backbones on three benchmarks.
\end{itemize}



\section{Related Work}

\paragraph{Novel-view reconstruction and geometric accuracy.}
3D Gaussian Splatting has been widely used for novel view synthesis, including feed-forward reconstruction from sparse images~\citep{chen2024mvsplat,zhang2025transplat,liu2025monosplat}. Dynamic methods extend static Gaussian Splatting~\citep{kerbl2023gaussian} by introducing deformation, trajectories, or time-varying attributes~\citep{luiten2024dynamic,wu2024four,yang2024deformable}, separating static and dynamic primitives~\citep{wu2025swift4d}, or reconstructing scenes from sparse camera observations~\citep{zhou2026c4d}. Surface reconstruction methods improve geometric accuracy by modifying Gaussian primitives and depth rendering during optimization~\citep{huang2024twodgs,chen2025pgsr}. Geometry-aware 4D generation methods predict consistent RGB-D sequences of future scenes~\citep{liu2026fourdgen}, including robotic manipulation tasks from LBM~\citep{tri2025lbm}. Novel-view depth accuracy has also been studied in MVSNeRF and FreeSplat~\citep{chen2021mvsnerf,wang2024freesplat}. Unlike these approaches, \method{} refines the rendered geometry of frozen reconstruction backbones without modifying their underlying representations. It can use depth predictions from the backbone itself or from a pretrained depth model such as Depth Anything~3~\citep{lin2026depthanything3}.

\paragraph{Multi-view geometric fusion and selective prediction.}
Reprojected observations have been used for visibility-aware image blending~\citep{hedman2018deepblending} and learned multi-sensor depth fusion~\citep{sandstrom2022senfunet}. Recent generative methods further exploit reprojected 3D information and multi-view consistency to improve generation quality~\citep{ren2025gen3c,yang2024viewfusion}. Selective prediction methods address unreliable predictions by determining when to accept or reject model outputs~\citep{geifman2017selective,jiang2026mail}. Different from these methods, \method{} focuses on selecting complementary geometric predictions to refine novel-view depth. Inspired by cost-sensitive learning~\citep{elkan2001foundations}, our selection mechanism considers the expected reduction in depth error rather than the probability of improvement alone.

\paragraph{Multi-view benchmarks.}
Existing multi-view benchmarks provide different forms of geometric supervision and vary in their capture settings and evaluation protocols. As summarized in Table~\ref{tab:rig-datasets}, RIG complements these benchmarks by providing synchronized real-world RGB-D observations and held-out depth measurements, enabling a unified evaluation of static and dynamic reconstruction models.

\begin{table}[t]
  \centering
  \footnotesize
  \setlength{\tabcolsep}{4.1pt}
\caption{Comparison of RIG with existing multi-view benchmarks.
\emph{Sync.\ cameras} denotes the number of synchronized cameras
(--- indicates moving or handheld cameras).
\emph{Held-out depth} indicates whether reference depth is available
for views excluded from reconstruction inputs.
\emph{Static/Dynamic} indicates whether published evaluations cover
static or dynamic reconstruction models.}
  \label{tab:rig-datasets}
  \begin{tabular}{lcccccc}
    \toprule
    Dataset & \rotatebox{90}{Real} & \rotatebox{90}{\begin{tabular}{@{}l@{}}Sync.\\cameras\end{tabular}} & Depth reference & \rotatebox{90}{\begin{tabular}{@{}l@{}}Held-out\\depth\end{tabular}} & \rotatebox{90}{Static} & \rotatebox{90}{Dynamic} \\
    \midrule
    DTU~\citep{jensen2014dtu} & \cmark & --- & Structured light & \cmark & \cmark & \xmark \\
    ScanNet++~\citep{yeshwanth2023scannetpp} & \cmark & --- & Laser scan & \cmark & \cmark & \xmark \\
    Neural 3D Video~\citep{li2022neural} & \cmark & 21 & \xmark & \xmark & \xmark & \cmark \\
    DyCheck iPhone~\citep{gao2022monocular} & \cmark & 1+2 & LiDAR & \xmark & \xmark & \cmark \\
    Panoptic Studio~\citep{joo2019panoptic} & \cmark & 521 & Time-of-flight & \cmark & \xmark & \xmark \\
    DexYCB~\citep{chao2021dexycb} & \cmark & 8 & Active stereo & \cmark & \xmark & \xmark \\
    LBM~\citep{tri2025lbm,liu2026fourdgen} & \xmark & 16 & Rendered & \cmark & \xmark & \cmark \\
    \midrule
    \rowcolor{gray!12} RIG (ours) & \cmark & 3 & Time-of-flight & \cmark & \cmark & \cmark \\
    \bottomrule
  \end{tabular}
\end{table}


\section{Methodology}
\method{} addresses geometric reliability in novel view synthesis by jointly assessing reconstruction predictions and complementary multi-view depth evidence to guide selective geometric correction. As illustrated in Figure~\ref{fig:schematic}, depth estimates from multiple source views are registered to the target view and aggregated through robust consensus to construct a candidate depth estimate and associated reliability cues. An admission mechanism then evaluates the candidate against the backbone prediction to determine where geometric correction is beneficial, enabling selective depth refinement while preserving reliable backbone predictions.

\subsection{Complementary Depth Errors}
\label{sec:method-attribution}

\begin{figure}[t]
  \centering
  \suppressfloats[t]
  \includegraphics[width=\linewidth]{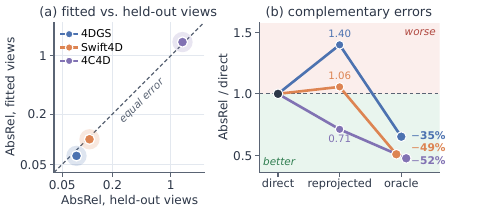}
  \caption{Depth-error attribution. (a)~AbsRel at held-out and fitted views. (b)~AbsRel normalized by the direct prediction, for depth reprojected from a fitted camera and a per-pixel oracle.}
  \label{fig:attribution}
\end{figure}

\paragraph{Attribution.}
Depth errors are not confined to unobserved viewpoints. Figure~\ref{fig:attribution}a shows that, across three dynamic Gaussian backbones, depth rendered at the twelve fitted cameras remains nearly as inaccurate as at held-out views, despite direct color supervision. These results establish that depth errors are not merely a consequence of novel-view extrapolation. A fundamental limitation lies in the depth rendering process: splatting computes depth as an opacity-weighted average of primitive depths along each ray, which need not correspond to a physical surface when primitives are dispersed in depth or partially transparent~\citep{huang2024twodgs,chen2025pgsr}.

Figure~\ref{fig:attribution}b evaluates the complementarity of depth predictions from the same frozen representation. AbsRel is normalized to the corresponding direct-rendering error at each held-out view. Depth reprojected from a fitted camera does not consistently improve upon direct rendering and can yield higher errors. In contrast, a per-pixel oracle selecting the estimate with lower ground-truth depth error reduces AbsRel by 35--52\% across the three backbones. This improvement demonstrates complementary error patterns: reprojected depth can provide more accurate local geometry even when its overall error is higher. This complementarity motivates selective correction based on the expected error reduction from replacing the direct prediction.

\subsection{Problem Formulation}
\label{sec:setting}

Let \(\mathcal C\) denote the set of calibrated cameras and \(\mathcal O\subset\mathcal C\) the source cameras available at the queried frame. For a target camera \(h\in\mathcal C\setminus\mathcal O\), a frozen backbone \(B\) renders color and depth:
\begin{equation}
  (I_B,D_B)=B(x_B),
\end{equation}
where \(x_B\) follows the backbone's prescribed input protocol, and \(I_B\) and \(D_B\) denote the rendered color image and depth, respectively. The backbone's weights and renderer remain unchanged.

Given the backbone predictions, source-view RGB observations, corresponding depth estimates, and camera calibration, our objective is to improve target-view depth accuracy while preserving rendering quality. We seek refined outputs \((\widehat I,\widehat D)\) without modifying the reconstruction backbone or accessing target-view observations.

\subsection{Multi-View Candidate Construction}
\label{sec:method-consensus}

\paragraph{Evidence sources.}
The evidence interface requires a depth estimate for each source camera that can be registered using known camera calibration. \methodself{} renders depth at the source cameras using the frozen backbone's scene representation, renderer, checkpoint, and depth readout. These estimates share the backbone's coordinate system and depth scale without requiring an external model. \methodda{} estimates depth from the same source RGB images using a frozen Depth Anything~3 model~\citep{lin2026depthanything3}, jointly conditioned on source-view poses with metric scale recovered from the known camera centres.

\paragraph{Calibrated reprojection and robust consensus.}
For each source camera \(c\in\mathcal O\), depth estimates \(Z_c\) are independently backprojected to 3D and reprojected into the target view \(h\) using known camera calibration. A hard z-buffer retains the nearest projected point at each target pixel \(u\), yielding registered depth \(D_c(u)\) and color \(I_c(u)\).

Following validity filtering, the registered estimates are aggregated in two passes. The first median establishes a reference depth for rejecting estimates whose relative deviation exceeds a tolerance \(\tau_D\). A second median over the remaining inlier set \(\mathcal V(u)\) yields the consensus candidate depth \(D_C\) and color image \(I_C\):
\begin{equation}
  D_C(u)=\operatorname*{median}_{c\in\mathcal V(u)}D_c(u),
  \qquad
  I_C(u)=\operatorname*{median}_{c\in\mathcal V(u)}I_c(u).
\end{equation}
Median consensus limits the influence of inconsistent sources. When two estimates are retained, the median equals the arithmetic mean. A validity mask \(\mathrm{cand}(u)\) indicates candidate availability. Inlier support and disagreement are mapped to a confidence score \(q(u)\) through a monotone function. These observable cues inform subsequent admission and color refinement.

\subsection{Expected-Gain Admission}
\label{sec:method-admission}

\paragraph{Expected error reduction.}
Let \(D^\star(u)\) denote the target-view reference depth and \(\phi(u)\) the observable features available to the admission head. At a pixel with a valid candidate, replacing the backbone depth with the candidate yields an absolute depth-error reduction of
\begin{equation}
  \Delta(u)=\big|D_B(u)-D^\star(u)\big|
            -\big|D_C(u)-D^\star(u)\big|.
  \label{eq:gain}
\end{equation}
Under an \(L_1\) objective, adopting the candidate minimizes expected absolute depth error when \(\mathbb E[\Delta(u)\mid\phi(u)]>0\).

This criterion differs from selecting candidates solely by their probability of improvement, which does not account for the magnitude of gains and losses. Candidate corrections may reduce error at most pixels yet increase expected error when occasional large failures outweigh frequent small gains. Our expected-gain admission accounts for this asymmetry. 

\paragraph{Cost-weighted training.}
The admission head is trained with binary labels \(\mathbf 1[\Delta(u)>0]\) and per-pixel weights \(|\Delta(u)|\), assigning greater importance to decisions with larger potential gains or losses. Let \(p=\sigma(\ell)\), where \(\ell\) is the predicted logit, and define \(\Delta_+=\max(\Delta,0)\) and \(\Delta_-=\max(-\Delta,0)\). The conditional weighted binary cross-entropy risk is
\begin{equation}
  L(p)=-\mathbb E\!\left[
    \Delta_+\log p+\Delta_-\log(1-p)
    \;\middle|\;\phi\right].
  \label{eq:cwbce}
\end{equation}
For \(\mathbb E[|\Delta|\mid\phi]>0\), the population minimizer is
\begin{equation}
  p^\star=
  \frac{\mathbb E[\Delta_+\mid\phi]}
       {\mathbb E[\Delta_+\mid\phi]+\mathbb E[\Delta_-\mid\phi]}.
  \label{eq:bayes}
\end{equation}
Since \(\mathbb E[\Delta\mid\phi]=\mathbb E[\Delta_+\mid\phi]-\mathbb E[\Delta_-\mid\phi]\), the condition \(p^\star>1/2\) holds exactly when \(\mathbb E[\Delta\mid\phi]>0\). Thus, at the population optimum, the zero-logit threshold aligns with the expected-gain criterion. At inference, a candidate is admitted only when it is valid and the predicted logit is positive:
\begin{equation}
  A(u)=\mathrm{cand}(u)\wedge[\ell(u)>0].
  \label{eq:admission}
\end{equation}

\paragraph{Admission head and inference.}
The admission head receives eighteen per-pixel channels organized into five groups. These comprise (i) backbone depth and local-structure cues, (ii) candidate depth, its discrepancy from the backbone prediction, and consensus reliability cues, (iii) a depth-agreement indicator, (iv) image position and the number of frames since the target view was last observed, and (v) geometric cues capturing surface orientation relative to the viewing ray and cross-view photometric consistency for both depth hypotheses. A small transformer with local-window attention maps these features to the admission logit \(\ell(u)\) at full resolution. The admission head is the only trainable component of \method{}.

\subsection{Selective Refinement}
\label{sec:method-safety}

\paragraph{Selective depth correction.}
The admission map selects between the candidate and backbone depth:
\begin{equation}
  D_{\mathrm{sel}}(u)=
  \begin{cases}
    D_C(u), & A(u)=1,\\
    D_B(u), & \text{otherwise.}
  \end{cases}
  \label{eq:accept}
\end{equation}
Admission requires registered evidence, so pixels without a valid candidate retain the backbone prediction. Where the backbone reports no surface, an admitted candidate is retained only if a previously available target-view observation confirms a surface at that pixel. Otherwise, the hole is preserved. Since no such observations are available under the evaluated held-out protocols, these holes remain unchanged. Let $D_H$ denote the depth map after this restriction.

To reduce seams at replacement boundaries, the log-depth correction \(\log D_H-\log D_B\) is smoothed near corrected pixels using only neighbors with consistent backbone depth and, where valid, consistent candidate depth. The result is the refined depth \(\widehat D\).

\paragraph{Constrained color refinement.}
\label{sec:method-color}
Color is refined through a bounded residual:
\begin{equation}
  \widehat I(u)=I_B(u)+\alpha(u)\big(I_C(u)-I_B(u)\big).
\end{equation}
The confidence-scaled weight \(\alpha(u)\) is capped and set to zero unless depth agreement, photometric consistency, and minimum support criteria are satisfied. Color refinement operates independently of the learned depth-admission map.


\section{Experiment Setup}

\subsection{Benchmarks and Backbone}
Evaluation covers two established benchmarks, DTU~\citep{jensen2014dtu} for feed-forward reconstruction and LBM~\citep{tri2025lbm} for dynamic reconstruction, alongside the proposed RIG benchmark for real-sensor evaluation. Six Gaussian reconstruction backbones are evaluated, comprising three feed-forward models and three dynamic models.

\noindent\textbf{DTU.}
DTU is a standard multi-view stereo benchmark with structured-light reference geometry. Three feed-forward Gaussian backbones are evaluated: MVSplat~\citep{chen2024mvsplat}, TranSplat~\citep{zhang2025transplat}, and MonoSplat~\citep{liu2025monosplat}.

\noindent\textbf{LBM.}
LBM provides simulated robotic manipulation tasks with multi-view RGB-D sequences released through 4DGen~\citep{liu2026fourdgen}. Three dynamic Gaussian backbones are evaluated: 4DGS~\citep{wu2024four}, Swift4D~\citep{wu2025swift4d}, and 4C4D~\citep{zhou2026c4d}.

\subsection{RIG: Real-World Benchmark}
\label{sec:rig}

\begin{figure}[t]
  \centering
  \includegraphics[width=\linewidth]{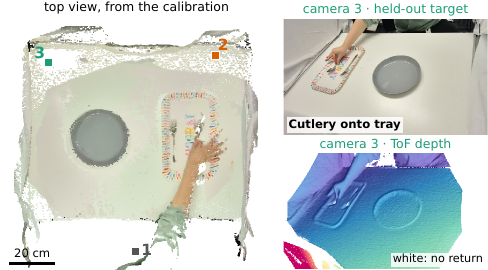}
  \caption{RIG benchmark and capture setup. \emph{Left:} Top view of fused ToF points from three synchronised RGB-D cameras. Two cameras provide source views, while the third serves as the held-out target. \emph{Right:} RGB and measured depth from a held-out camera, visualized as inverse depth (near red, far blue).}
  \vskip -0.1in
  \label{fig:rig}
\end{figure}

We introduce RIG, a multi-view RGB-D benchmark for evaluating the geometric accuracy of novel-view reconstruction. RIG uses three hardware-synchronized Orbbec Femto Bolt RGB-D cameras to capture tabletop object manipulation, with camera calibration performed using a ChArUco board (Figure~\ref{fig:rig}). 

One camera provides a fixed context view, while the two oblique cameras alternate as held-out targets. For each assignment, the other two cameras supply RGB inputs to the backbone and evidence for \method{}. Measured time-of-flight depth from the held-out camera serves as reference, excluding pixels without valid returns. Of 130 takes, 35 are reserved for admission-head training and validation, and 95 for evaluation. Across 95 held-out takes, 2,280 target views are evaluated per backbone.

\begin{table*}[!t]
  \centering
  \small
  \setlength{\tabcolsep}{10pt}
  \caption{Depth and appearance evaluation of three frozen feed-forward reconstruction backbones on DTU and RIG, comparing the original predictions with \methodself{} and \methodda{} refinement. Cand. reports AbsRel when all valid candidates are admitted. Shading highlights \method{} variants, and bold indicates improvements over the corresponding frozen backbone.}
  \label{tab:backbones}
  \begin{tabular}{lcrrrrrr}
    \toprule
    & & \multicolumn{2}{c}{Metric depth} & \multicolumn{4}{c}{Appearance} \\
    \cmidrule(lr){3-4} \cmidrule(lr){5-8}
    Method & Src. & Cand. $\downarrow$ & AbsRel $\downarrow$ & PSNR $\uparrow$ & SSIM-gray $\uparrow$ & SSIM-RGB $\uparrow$ & LPIPS $\downarrow$ \\
    \midrule
    \multicolumn{8}{l}{\emph{DTU, official cross-dataset test protocol}} \\
    MVSplat (ECCV 2024) & 2 & -- & 0.0440 & 12.71 & 0.3606 & 0.3584 & 0.4565 \\
    \rowcolor{gray!6} \quad + \methodself{} & 2 & 0.0425 & \textbf{0.0418} & \textbf{12.73} & \textbf{0.3702} & \textbf{0.3687} & \textbf{0.4547} \\
    \rowcolor{gray!12} \quad + \methodda{} & 2 & 0.0189 & \textbf{0.0180} & \textbf{12.74} & \textbf{0.3720} & \textbf{0.3709} & \textbf{0.4524} \\
    TranSplat (AAAI 2025) & 2 & -- & 0.0341 & 13.27 & 0.3883 & 0.3868 & 0.4129 \\
    \rowcolor{gray!6} \quad + \methodself{} & 2 & 0.0328 & \textbf{0.0327} & \textbf{13.30} & \textbf{0.3995} & \textbf{0.3984} & \textbf{0.4117} \\
    \rowcolor{gray!12} \quad + \methodda{} & 2 & 0.0181 & \textbf{0.0171} & \textbf{13.31} & \textbf{0.4010} & \textbf{0.4002} & \textbf{0.4104} \\
    MonoSplat (CVPR 2025) & 2 & -- & 0.0273 & 13.08 & 0.3761 & 0.3749 & 0.4084 \\
    \rowcolor{gray!6} \quad + \methodself{} & 2 & 0.0270 & \textbf{0.0266} & \textbf{13.11} & \textbf{0.3896} & \textbf{0.3889} & \textbf{0.4049} \\
    \rowcolor{gray!12} \quad + \methodda{} & 2 & 0.0184 & \textbf{0.0170} & \textbf{13.12} & \textbf{0.3889} & \textbf{0.3885} & \textbf{0.4051} \\
    \midrule
    \multicolumn{8}{l}{\emph{RIG, real-camera capture: 95 held-out takes, 2{,}280 targets per backbone}} \\
    MVSplat (ECCV 2024) & 2 & -- & 0.1626 & 11.53 & 0.4800 & 0.4779 & 0.6583 \\
    \rowcolor{gray!6} \quad + \methodself{} & 2 & 0.1645 & \textbf{0.1433} & 11.53 & 0.4794 & 0.4773 & \textbf{0.6567} \\
    \rowcolor{gray!12} \quad + \methodda{} & 2 & 0.0776 & \textbf{0.0718} & \textbf{11.54} & \textbf{0.4803} & \textbf{0.4785} & \textbf{0.6548} \\
    TranSplat (AAAI 2025) & 2 & -- & 0.1494 & 11.60 & 0.5096 & 0.5079 & 0.6499 \\
    \rowcolor{gray!6} \quad + \methodself{} & 2 & 0.1539 & \textbf{0.1369} & 11.60 & 0.5075 & 0.5061 & \textbf{0.6457} \\
    \rowcolor{gray!12} \quad + \methodda{} & 2 & 0.0742 & \textbf{0.0696} & \textbf{11.62} & 0.5088 & 0.5076 & \textbf{0.6440} \\
    MonoSplat (CVPR 2025) & 2 & -- & 0.1198 & 12.11 & 0.5445 & 0.5424 & 0.6254 \\
    \rowcolor{gray!6} \quad + \methodself{} & 2 & 0.1147 & \textbf{0.1074} & 12.11 & \textbf{0.5455} & \textbf{0.5435} & \textbf{0.6196} \\
    \rowcolor{gray!12} \quad + \methodda{} & 2 & 0.0672 & \textbf{0.0596} & \textbf{12.12} & \textbf{0.5467} & \textbf{0.5447} & \textbf{0.6187} \\
    \bottomrule
  \end{tabular}
\end{table*}

\subsection{Evaluation and Training Protocol}

\noindent\textbf{Backbone evaluation settings.}
On DTU, the three feed-forward backbones use their released RealEstate10K checkpoints without adaptation, following MVSplat's official cross-dataset protocol with two context views and four held-out targets per scene. On LBM, each dynamic backbone is fitted separately to each episode using its released code and published training schedule, with twelve source cameras and four held-out targets. On RIG, all three feed-forward backbones are evaluated at $256\times256$. Swift4D is fitted separately to each take and camera assignment using the same two context cameras.

\noindent\textbf{Paired evaluation.}
Each refined prediction is compared with its frozen backbone using the same task, target view, input cameras, checkpoint, and evaluation code. Both evidence variants use only the backbone's input cameras and require no additional observations or measured depth at inference (Section~\ref{sec:method-consensus}). The primary depth metric is absolute relative error (AbsRel), computed by averaging per-pixel absolute depth errors normalized by reference depth over valid pixels. Appearance metrics follow each backbone's evaluation code, with variants specified in the tables. Reference depth is used for admission-head supervision and evaluation, but is unavailable at inference.

\noindent\textbf{Admission training and fixed settings.}
Admission heads are trained separately for each benchmark and evidence source using the same architecture, objective, and zero-logit threshold. On DTU and RIG, feed-forward heads are trained on MVSplat outputs and transferred unchanged to TranSplat and MonoSplat. Each dynamic backbone has a separate head for each benchmark--evidence setting. All reconstruction backbones and the external depth model remain frozen during refinement.

On LBM, training and evaluation episodes are disjoint. On RIG, training and validation use only the 35 development takes. Evaluation targets are excluded from head training and parameter selection. Consensus and hole-handling parameters are selected on development data before evaluation and held fixed across benchmarks. The admission head is the only learned refinement component of \method{}.


\section{Experiment Result}
\label{sec:results}

\begin{table*}[t]
  \centering
  \small
  \setlength{\tabcolsep}{9.5pt}
  \caption{Depth and appearance evaluation of three frozen dynamic reconstruction backbones on LBM and RIG, comparing the original predictions with \methodself{} and \methodda{} refinement. Evidence variants and highlighting follow Table~\ref{tab:backbones}. $\delta_1$ denotes the fraction of valid pixels with corresponding depth ratio.}
  \label{tab:backbones4d}
  \begin{tabular}{lcrrrrrrr}
    \toprule
    & & \multicolumn{3}{c}{Metric depth} & \multicolumn{4}{c}{Appearance} \\
    \cmidrule(lr){3-5} \cmidrule(lr){6-9}
    Method & Src. & Cand. $\downarrow$ & AbsRel $\downarrow$ & $\delta_1$ $\uparrow$ & PSNR $\uparrow$ & SSIM $\uparrow$ & D-SSIM $\downarrow$ & LPIPS $\downarrow$ \\
    \midrule
    \multicolumn{9}{l}{\emph{LBM, four held-out views excluded from every backbone}} \\
    4DGS (CVPR 2024) & 12 & -- & 0.0744 & 0.9304 & 19.03 & 0.6779 & 0.0995 & 0.1880 \\
    \rowcolor{gray!6} \quad + \methodself{} & 12 & 0.0704 & \textbf{0.0582} & \textbf{0.9558} & \textbf{19.04} & \textbf{0.6798} & \textbf{0.0981} & \textbf{0.1866} \\
    \rowcolor{gray!12} \quad + \methodda{} & 12 & 0.1311 & \textbf{0.0391} & \textbf{0.9759} & 19.03 & \textbf{0.6781} & \textbf{0.0987} & 0.1882 \\
    Swift4D (ICLR 2025) & 12 & -- & 0.1064 & 0.8895 & 18.36 & 0.6601 & 0.1107 & 0.1853 \\
    \rowcolor{gray!6} \quad + \methodself{} & 12 & 0.0728 & \textbf{0.0580} & \textbf{0.9488} & \textbf{18.38} & \textbf{0.6632} & \textbf{0.1093} & \textbf{0.1844} \\
    \rowcolor{gray!12} \quad + \methodda{} & 12 & 0.1301 & \textbf{0.0384} & \textbf{0.9772} & \textbf{18.37} & \textbf{0.6607} & \textbf{0.1100} & 0.1860 \\
    4C4D (CVPR 2026) & 12 & -- & 1.3816 & 0.4159 & 20.87 & 0.7372 & 0.0711 & 0.2048 \\
    \rowcolor{gray!6} \quad + \methodself{} & 12 & 0.8617 & \textbf{0.7888} & \textbf{0.5456} & \textbf{20.89} & 0.7367 & \textbf{0.0705} & \textbf{0.1987} \\
    \rowcolor{gray!12} \quad + \methodda{} & 12 & 0.1392 & \textbf{0.0777} & \textbf{0.9633} & \textbf{20.88} & \textbf{0.7373} & \textbf{0.0704} & \textbf{0.2008} \\
    \midrule
    \multicolumn{9}{l}{\emph{RIG, real-camera capture: 95 held-out takes, 2{,}280 targets}} \\
    Swift4D (ICLR 2025) & 2 & -- & 0.1315 & 0.8030 & 12.10 & 0.5303 & 0.2348 & 0.4527 \\
    \rowcolor{gray!6} \quad + \methodself{} & 2 & 0.1227 & \textbf{0.1039} & \textbf{0.8786} & 12.10 & \textbf{0.5332} & \textbf{0.2334} & \textbf{0.4496} \\
    \rowcolor{gray!12} \quad + \methodda{} & 2 & 0.0805 & \textbf{0.0701} & \textbf{0.9120} & \textbf{12.11} & \textbf{0.5353} & \textbf{0.2324} & \textbf{0.4486} \\
    \bottomrule
  \end{tabular}
\end{table*}

\begin{figure*}[!t]
  \centering
  \includegraphics[width=\textwidth]{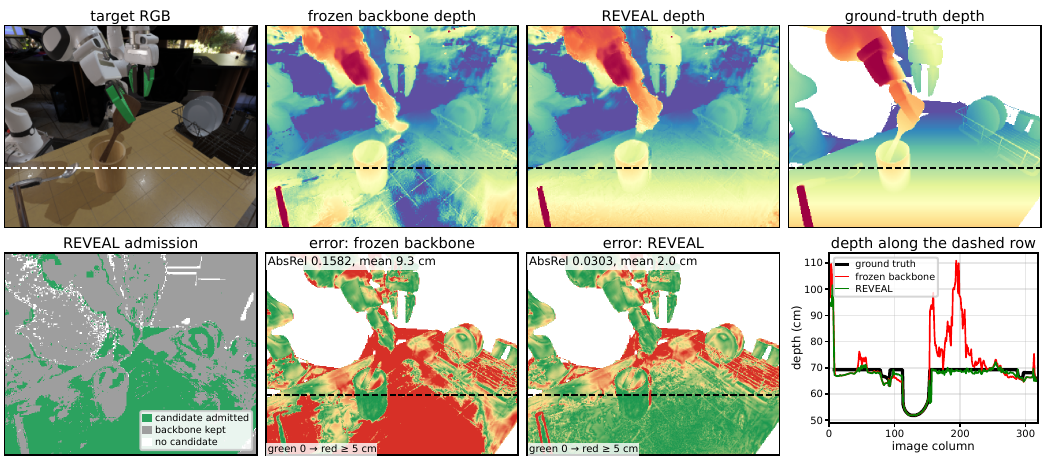}
  \caption{Depth refinement on an official held-out LBM view using a frozen dynamic backbone fitted to twelve source cameras. \emph{Top:} Target-view RGB and inverse-depth maps for the backbone, \method{}, and ground truth (near: red, far: blue, invalid: white). \emph{Bottom:} Admission map (green: candidate admitted, gray: backbone retained, white: no candidate), absolute depth-error maps for the backbone and \method{} (green: zero, red: $\geq 5$\,cm), and depth profiles along the dashed row.}
  \label{fig:holefill-real}
\end{figure*}

\subsection{Evaluation on DTU and LBM Benchmarks}
\label{sec:res-backbones}

Both evidence variants reduce AbsRel across all evaluated backbone--benchmark pairs (Tables~\ref{tab:backbones} and~\ref{tab:backbones4d}), and the reduction holds for individual targets and frames, not only on average: every scored LBM frame improves with \methodda{}, and at least $97.2\%$ with \methodself{} (Table~\otherref{tab:improved}{S1}). Improvements with \methodself{} directly support the attribution findings: querying the same frozen representation from the input cameras provides useful corrections to its target-view depth, without adding a depth estimator. The reduction is $4.1\%$ in mean AbsRel on DTU, with scene-level intervals excluding zero for every backbone, and $21.7$--$45.5\%$ across the dynamic backbones on LBM. These gains show that errors in the backbone's rendered depth leave room for correction using its own geometry. The corrections act on the errors that Figure~\ref{fig:attribution} attributes to depth rendering: where the depth is averaged off the surface, \method{} lowers AbsRel by $39$--$54\%$ with \methodself{} and $59$--$78\%$ with \methodda{}, and where no primitive lies at the surface, \methodda{} lowers it by $61$--$63\%$ (Appendix~\otherref{app:attribution}{F}).

The candidate columns reveal how evidence quality and selective admission contribute to refinement. On DTU, the pretrained depth model's registered candidate already accounts for much of the improvement, and admission reduces error further. On LBM, unconditional admission of that candidate increases error for 4DGS and Swift4D, yet \methodda{} reduces their AbsRel by $47.5\%$ and $63.9\%$, respectively. A candidate can therefore be less accurate overall while still providing valuable local corrections.

\begin{figure*}[!t]
  \centering
  \includegraphics[width=\textwidth]{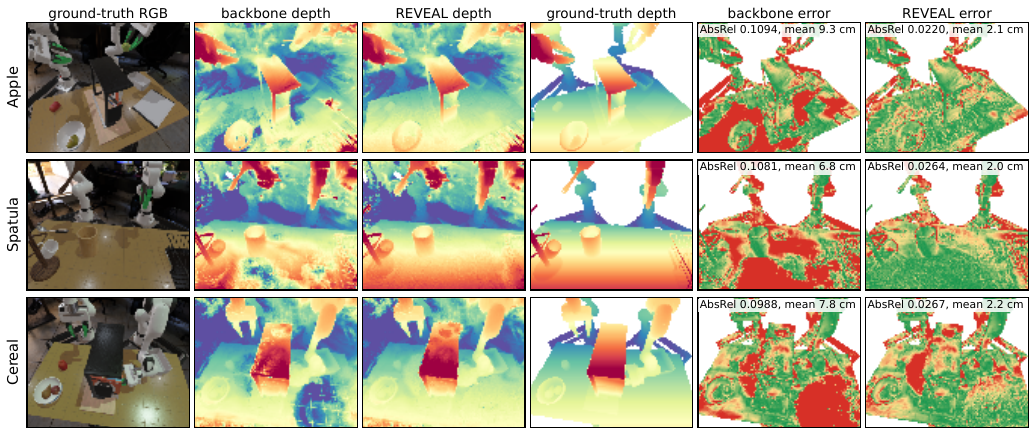}
  \caption{Depth refinement on held-out LBM views of the three tasks, one row each, for a frozen dynamic backbone fitted to twelve source cameras. Depth is drawn as inverse depth (near red, far blue), and error maps range from green at zero to red at $5$\,cm or above; each error panel gives its AbsRel and mean absolute error.}
  \label{fig:multitask}
\end{figure*}

The backbone comparison also demonstrates the separation between appearance and geometry. On LBM, 4C4D achieves the highest PSNR and SSIM among the evaluated backbones while producing the largest depth error. High image quality therefore provides limited evidence of accurate rendered depth. For Swift4D, the geometric improvement has a direct physical interpretation: \methodda{} reduces mean absolute depth error from $8.3$ to $3.1$\,cm and median error from $2.4$ to $1.4$\,cm, while appearance is preserved (Figures~\ref{fig:holefill-real} and~\ref{fig:multitask}). Figure~\ref{fig:multitask} further illustrates the improvements across three manipulation tasks, where refinement reduces broad regions of depth error while preserving the overall scene structure.

\subsection{Evaluation on RIG Benchmark}
\label{sec:res-real}

RIG evaluates whether these benefits persist with measured depth, moving hands, and real occlusions. With heads trained only on development takes, \methodself{} reduces the three static backbones' mean AbsRel by $10.2\%$, while \methodda{} reduces it by $53.4\%$. The corresponding improvements cover $98.8\%$ and $99.5\%$ of evaluated targets, showing that the aggregate reductions extend across the test set. On the same captures, Swift4D improves by $21.0\%$ with \methodself{} and $46.7\%$ with \methodda{}. The shared real-sensor protocol thus demonstrates effective refinement for both reconstruction families.

Selective admission remains important under these capture conditions. For MVSplat and TranSplat, unconditional use of the self-depth candidate increases error, whereas learned admission improves both backbones. This repeats the pattern observed with pretrained evidence on LBM: useful local corrections can survive substantial errors elsewhere in the candidate. The larger gains from \methodda{} show the benefit of stronger external evidence, while \methodself{} establishes that improvement does not require an additional depth estimator. The admission heads trained on MVSplat transfer unchanged to TranSplat and MonoSplat on both DTU and RIG, improving depth accuracy across all evaluated settings. This suggests that the comparative cues remain informative across different feed-forward reconstruction architectures, beyond the backbone used for admission-head training.

Depth refinement introduces only limited changes to rendered appearance. On DTU, all reported appearance metrics improve, with intervals excluding zero (Appendix~\otherref{app:appearance}{E}). On RIG, LPIPS improves for every backbone and evidence source, while Swift4D also improves in SSIM and D-SSIM. On LBM, PSNR is maintained or slightly improved, with occasional regressions in other appearance metrics. These results show that \method{} substantially improves depth accuracy while largely preserving rendered appearance. The admission head contains only $81K$ trainable parameters, and the complete refinement takes approximately two seconds per frame (Appendix~\otherref{app:cost}{D}).
\label{sec:res-cost}

\subsection{Component Analysis}
\label{sec:res-rule}

\begin{table}[!t]
  \centering
  \footnotesize
  \setlength{\tabcolsep}{12pt}
  \caption{Ablation of \method{} on Swift4D with \methodself{} evidence on LBM. Rows compare depth agreement, consensus, and learned admission with  reference depth.}
  \label{tab:ablation}
  \begin{tabular}{lcc}
    \toprule
    Component & AbsRel $\downarrow$ & $\delta_1$ $\uparrow$ \\
    \midrule
    Frozen backbone & 0.1064 & 0.8895 \\
    \quad + Depth agreement & 0.1034 & 0.8945 \\
    \quad + Consensus & 0.0976 & 0.9030 \\
    \quad + Probability rule & 0.0591 & 0.9474 \\
    \quad + Expected gain (\method{}) & \cellcolor{gray!12}0.0580 & \cellcolor{gray!12}0.9488 \\
    \midrule
    \emph{Per-pixel oracle (reference depth)} & 0.0514 & 0.9566 \\
    \bottomrule
  \end{tabular}
\end{table}

Table~\ref{tab:ablation} presents the contribution of each \method{} component using \methodself{} on Swift4D in LBM. Admitting depth from the nearest source camera where it agrees with the backbone reduces AbsRel by 2.9\%, while incorporating consensus across all source cameras brings a further 5.6\% reduction. However, disagreement alone cannot determine which estimate is more accurate, as it may result from either an incorrect candidate or an incorrect backbone prediction. Learning to select between the two estimates using reliability cues further reduces AbsRel by 39.4\%.

The expected-gain objective weights each training pixel by the magnitude of its potential error change. Compared with probability-based training, it further reduces AbsRel by 1.9\% using the same architecture, inputs, and training data, with improvements observed in every test episode. This shows the benefit of considering the magnitude of potential gains and losses.

The cost-weighted objective aligns the zero-logit decision threshold with positive expected error reduction at the population optimum. On DTU, this threshold achieves 97.4\% and 99.96\% of the maximum attainable gain across thresholds for \methodself{} and \methodda{}, respectively, compared with 57.3\% and 97.4\% when all valid candidates are admitted. The benefit of expected-gain training also extends to real-world data. On RIG, it outperforms probability-based training across all $95$ evaluation takes using admission heads with the same architecture and training data.


\section{Conclusion}

Depth rendered from different views of a frozen splatting representation exhibits complementary local errors that can be used to improve target-view geometry. \method{} constructs candidate geometry through calibrated reprojection and multi-view consensus, then uses comparative admission to select corrections expected to reduce depth error. The refinement operates without modifying reconstruction backbone. RIG provides a shared real-sensor evaluation protocol for static and dynamic reconstruction using depth measured by a held-out camera. Across six frozen backbones on public benchmarks and RIG, \method{} improves depth accuracy while maintaining comparable rendered appearance.


{
    \small
    \bibliographystyle{ieeenat_fullname}
    \bibliography{main}
}

\clearpage
\setcounter{page}{1}
\renewcommand{\thepage}{S\arabic{page}}
\renewcommand{\thefigure}{S\arabic{figure}}
\renewcommand{\thetable}{S\arabic{table}}
\renewcommand{\theequation}{S\arabic{equation}}
\setcounter{figure}{0}
\setcounter{table}{0}
\setcounter{equation}{0}
\renewcommand{\thepage}{S\arabic{page}}
\renewcommand{\thefigure}{S\arabic{figure}}
\renewcommand{\thetable}{S\arabic{table}}
\renewcommand{\theequation}{S\arabic{equation}}
\maketitlesupplementary
\renewcommand{\topfraction}{0.95}\renewcommand{\bottomfraction}{0.8}\renewcommand{\textfraction}{0.05}\renewcommand{\floatpagefraction}{0.85}
\renewcommand{\dbltopfraction}{0.95}\renewcommand{\dblfloatpagefraction}{0.85}
\setcounter{topnumber}{4}\setcounter{bottomnumber}{2}\setcounter{totalnumber}{6}\setcounter{dbltopnumber}{4}
\makeatletter\setlength{\@fptop}{0pt}\setlength{\@dblfptop}{0pt}\makeatother  

\appendix

\section{RIG Benchmark Construction and Protocol}
\label{app:rig}

\paragraph{Capture and calibration.}
RIG was captured using three hardware-synchronised Orbbec Femto Bolt RGB-D cameras fixed around a tabletop workspace (Figure~\otherref{fig:rig}{4} in the main paper). Each camera records RGB images at $1920\times1080$ resolution and registered time-of-flight depth in millimetres at 15 or 30 frames per second. ChArUco calibration achieves reprojection errors of $0.56$ and $0.24$\,px for the two cameras relative to the reference camera. An independent cross-camera consistency check shows median depth disagreements of $2.6$--$10.9$ mm across camera pairs and takes, measured on surfaces visible to both cameras.

\paragraph{Takes and evaluation protocol.}
RIG comprises 130 takes of tabletop object manipulation performed by an author who consented to recording and data release. The tasks involve placing a pen into a holder, a block into a drawer, a carrot into a bowl, a ball into a box, and cutlery onto a tray. These takes are split into 30 for admission-head training, 5 for validation, and 95 for evaluation. For each take, twelve uniformly sampled frames are scored under two camera assignments. Camera 1 serves as a fixed source view, while Cameras 2 and 3 alternate between the held-out target view and the second source view. Depth is evaluated where both predicted and held-out time-of-flight reference depths are valid. Confidence intervals are bootstrapped over takes.

\section{Computational Cost}
\label{app:cost}

All timings were measured on a single NVIDIA RTX 5080 GPU (16GB), which was also used for the LBM experiments. The admission head is the only trainable component of \method{}, containing $81K$ parameters and requiring a training time of $3$--$6$ minutes. Refining one LBM episode takes $72$--$82$\,s, approximately $2$\,s per target frame, including source-render loading, registration, consensus, admission, depth and color refinement, and scoring. \methodda{} additionally requires approximately $2$ minutes per episode for source-view depth estimation. Median backbone fitting times are $6$, $46$, and $101$ minutes per episode for Swift4D, 4DGS, and 4C4D, respectively. All backbones remain frozen during refinement.

\section{Depth Refinement Analysis}
\label{app:qualitative}

The depth accuracy metric $\delta_1$ in
Table~\otherref{tab:backbones4d}{3} of the main paper
denotes the fraction of valid pixels satisfying
$\max(\hat D/D^\star, D^\star/\hat D)<1.25$. Table~\ref{tab:improved} reports the percentage of evaluated targets or frames on which \method{} reduces AbsRel for each backbone and benchmark. On DTU, \methodself{} reduces mean AbsRel from $0.03515$ to $0.03370$ ($4.1\%$) and improves $94.8\%$ of the $3\times64$ held-out targets. Scene-level bootstrapping over the 16 test scenes yields reductions of $5.0\%$ for MVSplat ($3.5$--$7.1\%$) and $3.9\%$ for TranSplat ($2.6$--$5.9\%$), with improvements on all 16 scenes. MonoSplat achieves a $2.8\%$ reduction ($1.4$--$5.0\%$), improving on 15 of the 16 scenes.

\begin{figure*}[t]
  \centering
  \includegraphics[width=1\textwidth]{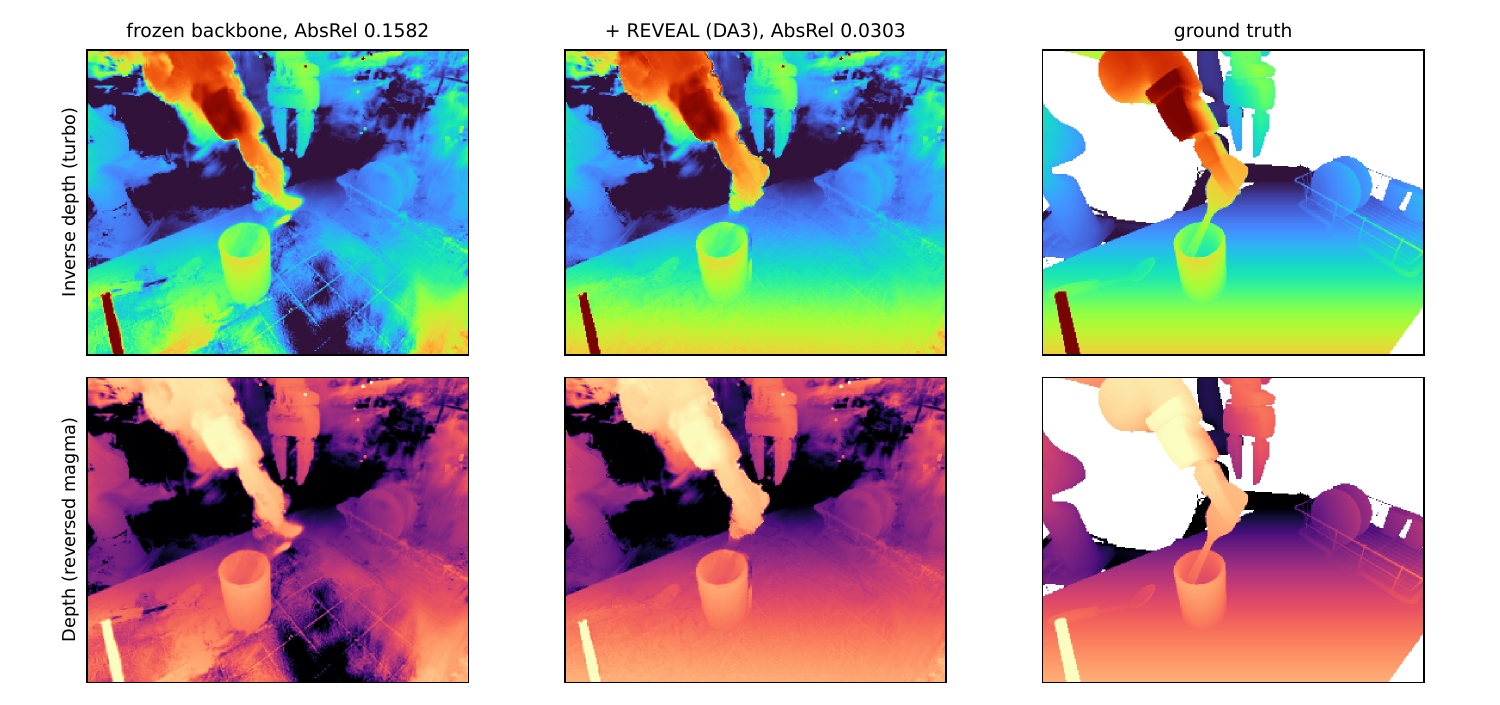}
  \caption{Alternative visualizations of the held-out LBM depth maps in
  Figure~\otherref{fig:holefill-real}{5} of the main paper.
  Rows show inverse depth (turbo) and depth (reversed magma).
  Each row uses the reference depth's 2nd--98th percentile range.
  White indicates invalid depth.}
  \label{fig:depth-styles}
\end{figure*}

Figure~\ref{fig:depth-styles} provides alternative visualizations of the held-out LBM example in Figure~\otherref{fig:holefill-real}{5} of the main paper. Both depth conventions show that the frozen backbone places large regions of the tabletop too far from the camera, while \methodda{} more closely follows the reference surface.

\begin{table}[t]
  \centering
  \small
  \setlength{\tabcolsep}{8.3pt}
  \caption{Percentage of evaluated targets (DTU, RIG) or frames
  (LBM) with lower AbsRel after \method{} refinement, compared
  with the frozen backbone. Results are shown for both evidence
  variants of Tables~\otherref{tab:backbones}{2}
  and~\otherref{tab:backbones4d}{3} in the main paper.}
  \label{tab:improved}
  \begin{tabular}{lrrrr}
    \toprule
    & \multicolumn{2}{c}{DTU} & \multicolumn{2}{c}{RIG} \\
    \cmidrule(lr){2-3} \cmidrule(lr){4-5}
    Static & Self-Depth & DA3 & Self-Depth & DA3 \\
    \midrule
    MVSplat & 98.4 & 100 & 99.0 & 99.6 \\
    TranSplat & 100 & 100 & 97.7 & 99.7 \\
    MonoSplat & 85.9 & 95.3 & 99.7 & 99.3 \\
    \midrule
    & \multicolumn{2}{c}{LBM} & \multicolumn{2}{c}{RIG} \\
    \cmidrule(lr){2-3} \cmidrule(lr){4-5}
    Dynamic & Self-Depth & DA3 & Self-Depth & DA3 \\
    \midrule
    4DGS & 97.2 & 100 & -- & -- \\
    Swift4D & 100 & 100 & 98.9 & 99.5 \\
    4C4D & 100 & 100 & -- & -- \\
    \bottomrule
  \end{tabular}
\end{table}

\paragraph{Robustness to limited depth evidence.}
Table~\ref{tab:nearest} evaluates \methodself{} on LBM using depth evidence from only the nearest source camera, without retraining the admission heads. AbsRel decreases by $13.5$--$30.3\%$ across the three dynamic backbones, with improvements on at least $95\%$ of frames and across all reported appearance metrics. Admitting all single-view candidates yields smaller gains and even increases error for 4DGS, highlighting the importance of selective admission. Thus, effective depth refinement does not require dense multi-view depth evidence, and its gains do not depend on a dense camera set.

\begin{table*}[t]
  \centering
  \small
  \setlength{\tabcolsep}{9.2pt}
  \caption{Single-view depth-evidence ablation of \methodself{} on held-out LBM views. Each backbone is fitted using twelve source cameras, but refinement uses depth evidence from only the nearest source camera. Admission heads remain unchanged, and photometric cues use all source views. Src. denotes the number of cameras used for backbone fitting (baseline) or depth evidence (refinement). Cand. reports AbsRel when all valid candidates are admitted. Bold indicates improvements over the corresponding frozen backbone.}
  \label{tab:nearest}
  \begin{tabular}{lcrrrrrrr}
    \toprule
    & & \multicolumn{3}{c}{Metric depth} & \multicolumn{4}{c}{Appearance} \\
    \cmidrule(lr){3-5} \cmidrule(lr){6-9}
    Method & Src. & Cand. $\downarrow$ & AbsRel $\downarrow$ & $\delta_1$ $\uparrow$ & PSNR $\uparrow$ & SSIM $\uparrow$ & D-SSIM $\downarrow$ & LPIPS $\downarrow$ \\
    \midrule
    \multicolumn{9}{l}{\emph{LBM, official test views held out from every backbone}} \\
    4DGS (CVPR 2024) & 12 & -- & 0.0744 & 0.9304 & 19.03 & 0.6779 & 0.0995 & 0.1880 \\
    \rowcolor{gray!6} \quad + \methodself{} & 1 & 0.0752 & \textbf{0.0643} & \textbf{0.9466} & \textbf{19.05} & \textbf{0.6819} & \textbf{0.0979} & \textbf{0.1854} \\
    Swift4D (ICLR 2025) & 12 & -- & 0.1064 & 0.8895 & 18.36 & 0.6601 & 0.1107 & 0.1853 \\
    \rowcolor{gray!6} \quad + \methodself{} & 1 & 0.0970 & \textbf{0.0816} & \textbf{0.9189} & \textbf{18.39} & \textbf{0.6653} & \textbf{0.1091} & \textbf{0.1830} \\
    4C4D (CVPR 2026) & 12 & -- & 1.3816 & 0.4159 & 20.87 & 0.7372 & 0.0711 & 0.2048 \\
    \rowcolor{gray!6} \quad + \methodself{} & 1 & 1.1162 & \textbf{0.9635} & \textbf{0.5124} & \textbf{20.91} & \textbf{0.7387} & \textbf{0.0698} & \textbf{0.1966} \\
    \bottomrule
  \end{tabular}
\end{table*}

\section{Admission Decision Analysis}
\label{app:learn}

On DTU, admission heads are trained on scans 2--7, 9, and 10 and validated on scans 11--14, all disjoint from the 16 test scenes.

\begin{figure}[!tbp]
  \centering
  \includegraphics[width=\linewidth]{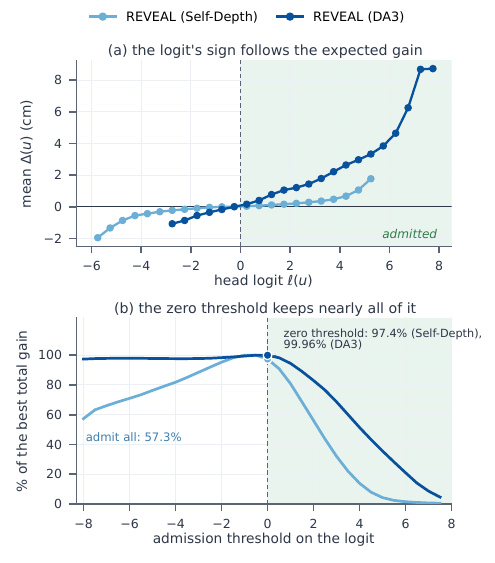}
  \caption{Expected-gain admission on held-out DTU views across
  three feed-forward backbones. (a) Mean depth-error reduction
  $\Delta(u)$ across logit bins containing at least $0.1\%$
  of pixels. (b) Total error reduction across admission thresholds,
  normalized by the maximum achievable gain. Dots mark the
  deployed zero threshold.}
  \label{fig:calibration}
\end{figure}

Figure~\ref{fig:calibration} shows that mean depth-error reduction increases with the admission logit and changes sign near zero. The deployed zero threshold achieves near-maximal total gain across thresholds, supporting the expected-gain admission criterion.

On RIG with \methodself{}, expected-gain training reduces AbsRel from $0.1446$ to $0.1433$ for MVSplat, $0.1382$ to $0.1369$ for TranSplat, and $0.1078$ to $0.1074$ for MonoSplat, compared with uniform-weight BCE under identical training settings.

\section{Appearance Preservation Analysis}
\label{app:appearance}

Table~\ref{tab:appearance-ci} reports relative changes in appearance metrics after applying \methodself{} on DTU, with 95\% cluster-bootstrap confidence intervals over all test scenes. All three backbones show improvements across PSNR, SSIM-gray, SSIM-RGB, and LPIPS, with all confidence intervals excluding zero. These results demonstrate that depth refinement not only improves geometric accuracy but also yields consistent gains in appearance quality.


\begin{table*}[t]
  \centering
  \small
  \setlength{\tabcolsep}{11.3pt}
  \caption{Relative change (\%) of each appearance metric with
  \methodself{} on DTU, compared with the frozen backbones.
  Brackets indicate 95\% scene-level bootstrap confidence
  intervals (20,000 resamples).}
  \label{tab:appearance-ci}
  \begin{tabular}{lrrrr}
    \toprule
    Backbone & PSNR $\uparrow$ & SSIM-gray $\uparrow$ & SSIM-RGB $\uparrow$ & LPIPS $\downarrow$ \\
    \midrule
    MVSplat & $+0.15$ [0.09, 0.21] & $+2.68$ [2.07, 3.30] & $+2.87$ [2.21, 3.56] & $-0.40$ [$-0.64$, $-0.17$] \\
    TranSplat & $+0.18$ [0.11, 0.26] & $+2.87$ [2.33, 3.43] & $+2.99$ [2.43, 3.55] & $-0.27$ [$-0.52$, $-0.05$] \\
    MonoSplat & $+0.23$ [0.16, 0.31] & $+3.59$ [2.93, 4.27] & $+3.74$ [3.00, 4.51] & $-0.86$ [$-1.30$, $-0.47$] \\
    \midrule
    Three-backbone mean & $\mathbf{+0.19}$ [0.15, 0.23] & $\mathbf{+3.05}$ [2.67, 3.42] & $\mathbf{+3.20}$ [2.80, 3.61] & $\mathbf{-0.50}$ [$-0.70$, $-0.32$] \\
    \bottomrule
  \end{tabular}
\end{table*}

\section{Depth-Error Attribution Analysis}
\label{app:attribution}

\begin{figure}[!tbp]
  \centering
  \includegraphics[width=\linewidth]{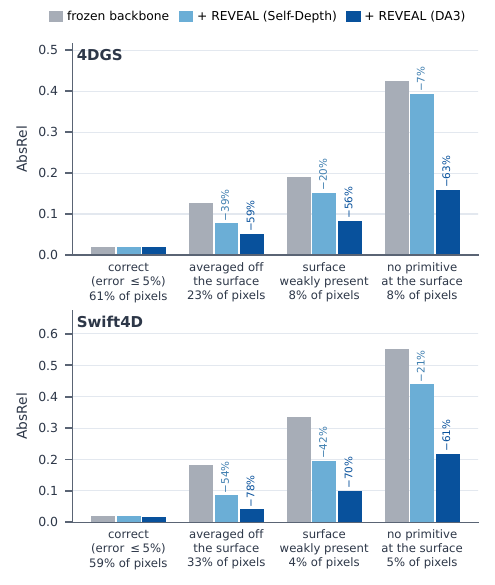}
  \caption{AbsRel by depth-error class on held-out LBM views from the local refits in Table~\ref{tab:attribution-full}. Pixels with relative depth error at most $5\%$ are considered correct. Incorrect pixels are grouped by the fraction of Gaussian compositing weight within $5\%$ of the reference depth: at least $25\%$ (averaged off the surface), $5$--$25\%$ (surface weakly present), and below $5\%$ (little surface support). Percentages below each class indicate its share of evaluated pixels, while bar annotations show the relative change in AbsRel from the frozen backbone.}
  \label{fig:errclass}
\end{figure}

Table~\ref{tab:attribution-full} extends the depth-error analysis in Figure~\otherref{fig:attribution}{3} of the main paper. It examines depth scale, error distribution, Gaussian depth spread, and complementary multi-view predictions. Unmarked rows use the original evaluated reconstructions, while $\dagger$ denotes additional diagnostics from local refits. Errors are evaluated where both prediction and reference depth are valid.

\paragraph{Depth accuracy and scale.}
Depth errors remain substantial even at the fitted views, indicating that inaccuracies cannot be explained by novel-view extrapolation alone. Correcting each frame's median predicted-to-reference depth ratio reduces AbsRel by only $3.9\%$ for 4DGS and $1.4\%$ for Swift4D. For 4C4D, the median ratio is $1.56$, indicating substantial depth overestimation. Although rescaling reduces its error by $46\%$, AbsRel remains $0.74$. These results show that global scale correction alone cannot resolve the depth errors.

\paragraph{Error distribution.}
Pixels within two pixels of a reference depth edge account for $15\%$ of evaluated pixels but contribute only $18\%$, $13\%$, and $6\%$ of the error for 4DGS, Swift4D, and 4C4D, respectively. Semi-transparent pixels (opacity below $0.99$) account for $3\%$, $5\%$, and $81\%$ of pixels, contributing $12\%$, $20\%$, and $99\%$ of the corresponding errors. Predictions lying more than $5\%$ behind or in front of every reference depth in a $5\times5$ neighbourhood account for $70\%$, $81\%$, and $99\%$ of total error. Thus, depth inaccuracies are not confined to occlusion boundaries or local depth discontinuities.

\paragraph{Gaussian depth spread and readouts.}
For the local refits ($\dagger$), Gaussian view depths $z$, $z^2$, $z^{-8}$, and $z^8$ are composited as auxiliary channels to measure relative depth spread and alternative depth readouts. Pixels with depth spread exceeding $10\%$ account for $42$--$44\%$ of pixels but contribute $74$--$81\%$ of error. For $51\%$ and $66\%$ of incorrect pixels, the reference depth is bracketed, within $5\%$, by the nearest- and farthest-weighted readouts $\mathbb{E}[z^{-8}]^{-1/8}$ and $\mathbb{E}[z^8]^{1/8}$. This indicates that contributing Gaussian depths can span the true surface even when their opacity-weighted average is inaccurate. Neither alternative readout improves upon the original depth estimate, motivating the use of complementary views rather than a fixed replacement readout.

\paragraph{Complementary multi-view depth.}
Each fitted-view depth is independently registered to the held-out view using the procedure in Section~\otherref{sec:method-consensus}{3.3} of the main paper. A single reprojected view has $1.40$ and $1.06$ times the direct-rendering AbsRel for 4DGS and Swift4D, respectively, yet provides a more accurate estimate at $41$--$64\%$ of jointly valid pixels. A per-pixel oracle reduces AbsRel to $0.48$--$0.65$ times the backbone error. Median consensus across all twelve source views improves candidate accuracy, achieving $0.56$--$0.97$ times the backbone AbsRel, while the corresponding oracle reaches $0.46$--$0.64$. These results demonstrate the complementary roles of multi-view consensus and selective admission. Table~\ref{tab:nearest} further shows that restricting depth evidence to the nearest source camera still reduces AbsRel by $13.5$--$30.3\%$, with photometric cues retained from all source views.

\paragraph{Errors corrected by \method{}.}
Figure~\ref{fig:errclass} shows that \method{} preserves accurate pixels, with AbsRel differing from the backbone by at most $0.001$, while reducing error across all incorrect classes. Where depth is averaged off the surface, \methodself{} reduces AbsRel by $39$--$54\%$ and \methodda{} by $59$--$78\%$. Even for pixels where less than $5\%$ of the compositing weight is concentrated near the true surface, \methodda{} reduces AbsRel by $61$--$63\%$. These findings show that source-view evidence can correct errors in the backbone's depth readout, while external depth evidence can also improve regions with little Gaussian support near the reference surface.

\begin{table}[t]
  \centering
  \footnotesize
  \setlength{\tabcolsep}{3.6pt}
  \caption{Depth-error attribution for three frozen dynamic
backbones on held-out LBM views. Unmarked rows use the
evaluated reconstructions, while $\dagger$ denotes
diagnostics from local refits of 4DGS and Swift4D.
Entries labelled ``pixels/error'' report the percentage
of evaluated pixels and their contribution to total
depth error, respectively. Predictions behind or in
front of nearby surfaces differ by more than $5\%$
from every reference depth in a $5\times5$ neighbourhood.
A surface is considered bracketed when the reference
depth lies between the nearest- and farthest-weighted
readouts within $5\%$. Reprojection results are AbsRel
ratios relative to the backbone on jointly valid
pixels. Per-pixel oracles use reference depth to select
the lower-error estimate.}
  \label{tab:attribution-full}
  \begin{tabular}{@{}lrrr@{}}
    \toprule
     & 4DGS & Swift4D & 4C4D \\
    \midrule
    \multicolumn{4}{@{}l}{\emph{Depth accuracy and scale}} \\
    Held-out AbsRel $\downarrow$ & 0.0744 & 0.1064 & 1.3816 \\
    Fitted-view AbsRel $\downarrow$ & 0.0633 & 0.1000 & 1.4579 \\
    Median pred./ref. depth ratio $\to 1$ & 0.98 & 0.99 & 1.56 \\
    AbsRel after rescaling $\downarrow$ & 0.0715 & 0.1049 & 0.7398 \\
    \midrule
    \multicolumn{4}{@{}l}{\emph{Error distribution}} \\
    Semi-transparent pixels: pixels/error & 3/12\% & 5/20\% & 81/99\% \\
    Depth edges: pixels/error & 15/18\% & 15/13\% & 15/6\% \\
    Error share behind nearby surfaces & 38\% & 59\% & 99\% \\
    Error share in front of nearby surfaces & 32\% & 22\% & 0\% \\
    \midrule
    \multicolumn{4}{@{}l}{\emph{Gaussian depth spread and readouts}} \\
    $^\dagger$Held-out AbsRel of the refits $\downarrow$ & 0.0868 & 0.0871 & -- \\
    $^\dagger$Contributors spread $>10\%$: pixels/error & 42/74\% & 44/81\% & -- \\
    $^\dagger$Wrong pixels, surface among contributors & 51\% & 66\% & -- \\
    $^\dagger$Nearest-weighted readout AbsRel $\downarrow$ & 0.1828 & 0.1453 & -- \\
    $^\dagger$Farthest-weighted readout AbsRel $\downarrow$ & 0.2246 & 0.4048 & -- \\
    \midrule
    \multicolumn{4}{@{}l}{\emph{Complementary multi-view depth}} \\
    Single-view reprojection / backbone $\downarrow$ & 1.40 & 1.06 & 0.71 \\
    \quad pixels where it is closer & 42\% & 41\% & 64\% \\
    \quad per-pixel oracle / backbone $\downarrow$ & 0.65 & 0.51 & 0.48 \\
    Twelve-view median / backbone $\downarrow$ & 0.97 & 0.68 & 0.56 \\
    \quad per-pixel oracle / backbone $\downarrow$ & 0.64 & 0.46 & 0.47 \\
    \bottomrule
  \end{tabular}
\end{table}

\end{document}